%% file: sample-sigconf-authordraft.tex
\documentclass[sigconf]{acmart}
\usepackage{array}
\usepackage{multirow}
\usepackage{balance}

\definecolor{pastelblue}{RGB}{204,229,255}
\definecolor{pastelgreen}{RGB}{204,255,229}
\definecolor{pastelyellow}{RGB}{255,255,204}
\definecolor{pastelpink}{RGB}{255,204,229}
\definecolor{pastelpurple}{RGB}{229,204,255}

\AtBeginDocument{%
  }

\copyrightyear{2025}
\acmYear{2025}
\setcopyright{acmlicensed}
\setcctype{by-nc-nd}
\acmConference[MM '25]{Proceedings of the 33rd ACM International Conference on Multimedia}{October 27--31, 2025}{Dublin, Ireland}
\acmBooktitle{Proceedings of the 33rd ACM International Conference on Multimedia (MM '25), October 27--31, 2025, Dublin, Ireland}
\acmDOI{10.1145/3746027.3755060}
\acmISBN{979-8-4007-2035-2/2025/10}

\acmSubmissionID{1885}

\begin{teaserfigure}
  \centering
  \includegraphics[width=0.8\textwidth]{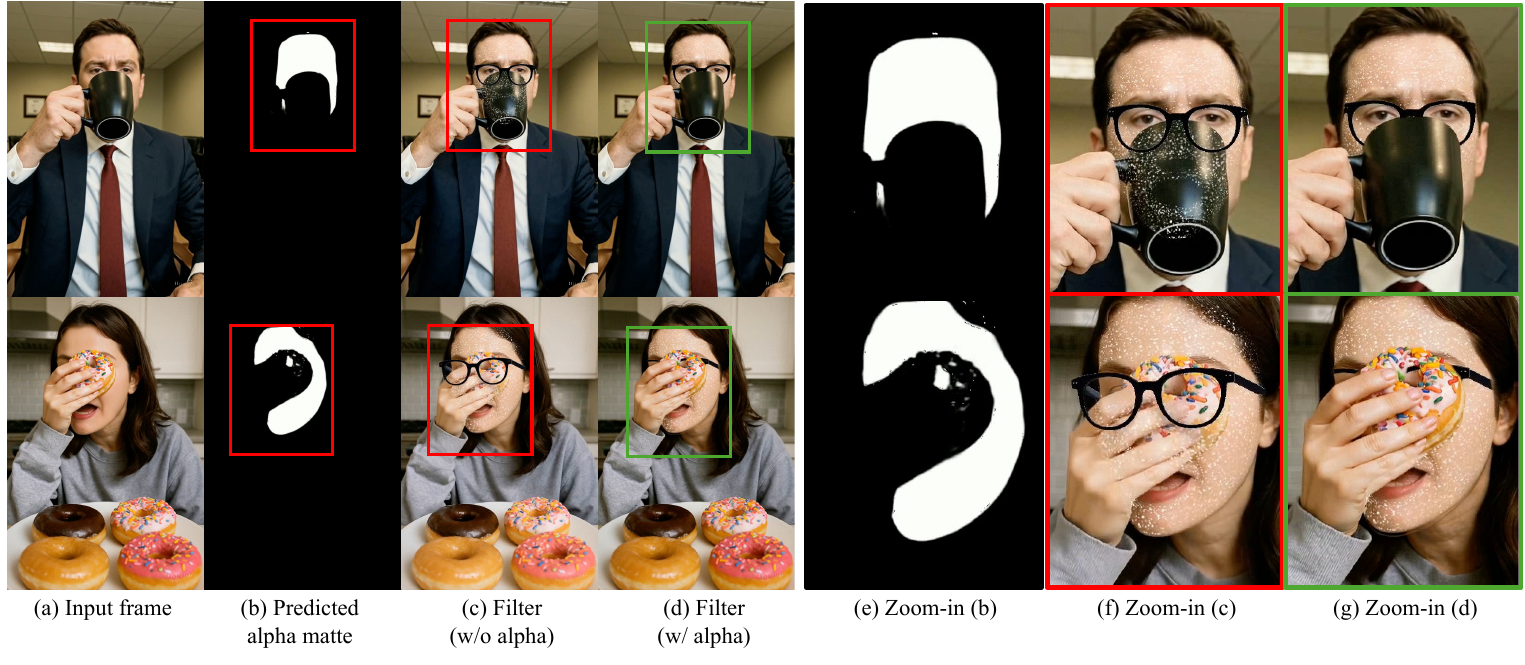}
  \vspace{-0.3cm}
  \caption{Application of face matting for occlusion-aware face filtering. (a) shows the original input image with partial occlusions. (b) presents the alpha matte predicted by our FaceMat framework, which separates occluding elements (e.g., hand, microphone) from the facial region. (c, d) compare visual filter results without and with alpha matte, respectively. Without matting (c), the filter is incorrectly applied to occluders, leading to unnatural results. In contrast, (d) shows alpha-guided compositing using the output from (b), where the filter is confined to the facial area. (e, f) show zoomed-in versions for clearer comparison.}
  \label{fig:teaser}
\end{teaserfigure}

\begin{document}

\title{Uncertainty-Guided Face Matting for Occlusion-Aware Face Transformation}

\author{Hyebin Cho}
\affiliation{%
  \institution{Korea Advanced Institute of Science \& Technology} 
  \department{School of Electrical Engineering}
  \city{Daejeon}
  \country{Republic of Korea}
  }
\email{hyebin.cho@kaist.ac.kr} 


\author{Jaehyup Lee}
\authornote{Corresponding author}
\affiliation{%
  \institution{Kyungpook National University}
    \department{School of Computer Science and Engineering}
  \city{Daegu}
  \country{Republic of Korea}}
\email{jaehyuplee@knu.ac.kr}

\renewcommand{\shortauthors}{Hyebin Cho and Jaehyup Lee}


\begin{abstract}
Face filters have become a key element of short-form video content, enabling a wide array of visual effects such as stylization and face swapping. However, their performance often degrades in the presence of occlusions, where objects like hands, hair, or accessories obscure the face. To address this limitation, we introduce the novel task of face matting, which estimates fine-grained alpha mattes to separate occluding elements from facial regions. We further present FaceMat, a trimap-free, uncertainty-aware framework that predicts high-quality alpha mattes under complex occlusions. Our approach leverages a two-stage training pipeline: a teacher model is trained to jointly estimate alpha mattes and per-pixel uncertainty using a negative log-likelihood (NLL) loss, and this uncertainty is then used to guide the student model through spatially adaptive knowledge distillation. This formulation enables the student to focus on ambiguous or occluded regions, improving generalization and preserving semantic consistency. Unlike previous approaches that rely on trimaps or segmentation masks, our framework requires no auxiliary inputs making it well-suited for real-time applications. In addition, we reformulate the matting objective by explicitly treating skin as foreground and occlusions as background, enabling clearer compositing strategies. To support this task, we newly constructed CelebAMat, a large-scale synthetic dataset specifically designed for occlusion-aware face matting. Extensive experiments show that FaceMat outperforms state-of-the-art methods across multiple benchmarks, enhancing the visual quality and robustness of face filters in real-world, unconstrained video scenarios. The source code and CelebAMat dataset are available at \textit{https://github.com/hyebin-c/FaceMat.git}.

\end{abstract}




\maketitle

\section{Introduction}
\indent


\indent

With the increasing popularity of short-form content on platforms like TikTok, Instagram, and YouTube Shorts, face filtering has emerged as a key feature for enhancing user engagement. These filters perform facial region detection and apply various visual effects such as stylization, overlays, and face swapping to enhance visual storytelling and provide personalized content. 

\begin{figure}[t]
    \includegraphics[width=\linewidth]{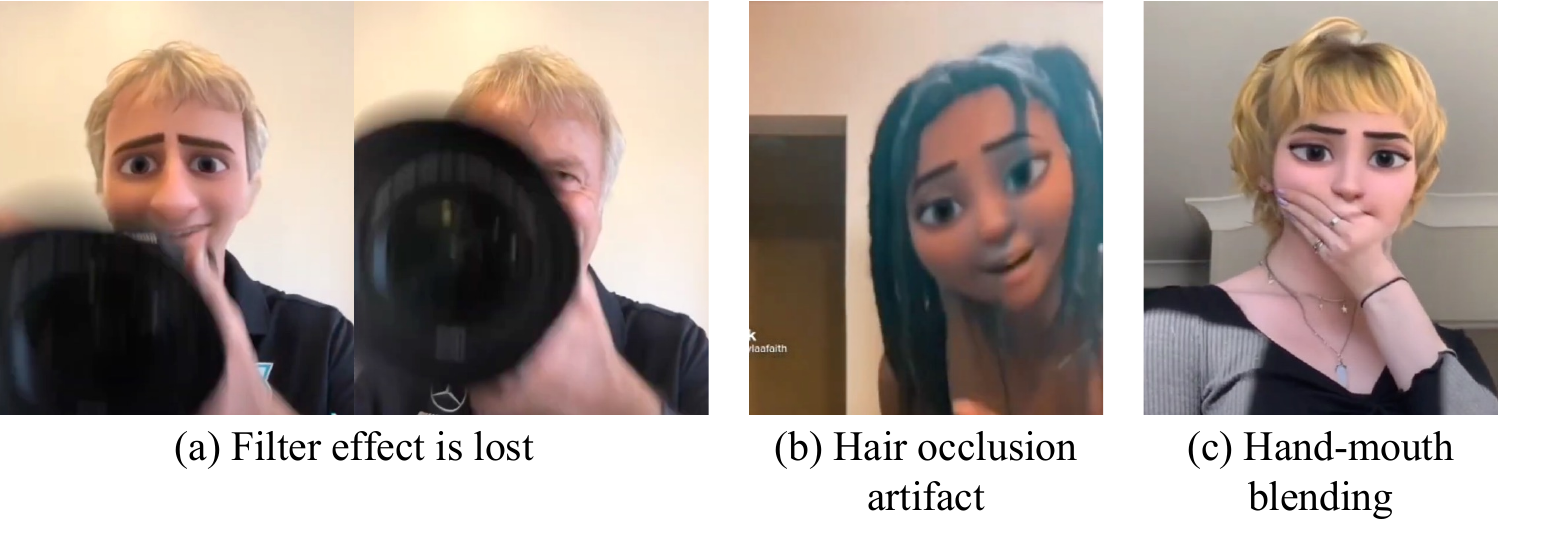}
    \vspace{-0.4cm}
    \caption{Limitation of face manipulation under occlusion. The effectiveness of face-related applications may degrade significantly due to failures in face recognition or a lack of occlusion-aware design in manipulation techniques.}
\label{fig:application_limitation}
\end{figure}

However, as shown in Fig. \ref{fig:application_limitation}, existing face filtering techniques often degrade under real-world conditions, particularly in the presence of occlusions caused by motion-blurred hands, accessories, or hair. Such occlusions often cause unnatural artifacts, as conventional segmentation methods relying on binary masks fail to capture fine-grained transparency. Consequently, they can not properly handle subtle transitions between foreground and background, resulting in degraded filter quality.

Image matting addresses this limitation by estimating a per-pixel alpha matte that models soft transitions between foreground and background. A pixel $I_i$ in the image can be represented as:

\begin{equation} \label{eq:1} I_i = \alpha_i F_i + (1-\alpha_i) B_i, \quad \alpha \in [0,1] \end{equation}

where $F_i$, $B_i$ represent the foreground and background color components of the $I_i$, and $\alpha_i$ denotes the corresponding alpha value.  

While image matting provides a precise representation, predicting alpha mattes from a single image remains an ill-posed problem. Most of the previous work relies heavily on auxiliary priors such as trimaps \cite{DIM, aematter, matteformer}, binary masks \cite{mgmatting, wildmatting}, or background images \cite{background} to constrain the solution. However, such inputs are difficult to be obtained in real-time video settings, making them impractical for dynamic face filtering applications.

Recent advances in trimap-free matting aim to eliminate the dependence on auxiliary inputs by training data-driven models in specific domains such as human portraits \cite{modnet, RVM}. However, these models fail to operate reliably under challenging conditions such as occlusions, motion blur, and acquisition noise, which are common in short-form video content. 

To address these challenges, we propose FaceMat, a robust and trimap-free face matting framework designed for occlusion-aware visual effects. At the core of our method is an uncertainty-aware knowledge distillation strategy. We first train a teacher model to jointly predict alpha mattes and per-pixel uncertainty using a negative log-likelihood (NLL) loss. The resulting uncertainty maps, which reflect the teacher model's confidence, are then used to adaptively adjust the distillation temperature across spatial locations. This allows the student model to focus its learning on ambiguous or occluded regions, resulting in improved generalization and robustness.




We further reinterpret the matting task by treating facial skin as background and occlusions such as hands or objects as foreground. Based on this definition, our proposed FaceMat operates in four stages: (1) Occlusion matting, where an alpha matte is predicted to isolate occluding elements from the facial region; (2) Face completion, where an inpainting module can optionally reconstruct occluded facial areas to obtain a clean face; (3) Face transformation, which applies visual effects such as swapping or stylization to the completed face; and (4) Compositing, where the transformed face is blended with the original occlusion using the predicted alpha matte to ensure visual consistency. 
This design enables natural and realistic face filtering under complex occlusions while preserving realistic appearances. Our contributions are summarized as follows:

\begin{itemize}
    \item We define \textit{face matting} as a new task that explicitly separates facial occlusions from skin regions, reformulating the foreground-background relationship.
    \item We introduce \textbf{FaceMat}, a trimap-free matting framework guided by per-pixel uncertainty to enable locally adaptive learning via knowledge distillation.
    \item We propose a multi-stage pipeline that integrates matting, inpainting, transformation, and compositing for high-quality filtering under occlusion.
    \item We construct a synthetic dataset, \textbf{CelebAMat}, designed for face matting and demonstrate that FaceMat achieves state-of-the-art results in challenging video matting applications with frequent facial occlusions.
\end{itemize}

\section{Related Work}
 
\subsection{Image and Video Matting}
\indent
Image matting is an inherently ill-posed problem that often requires strong priors to produce reliable alpha mattes. One of the most commonly used priors is the trimap, which divides the image into foreground, background, and unknown regions in the corresponding context. By providing explicit spatial constraints, trimaps significantly enhance matting accuracy and stability. Consequently, a variety of trimap-based approaches \cite{DIM, alphagan, matteformer, diffusion} have been proposed, leveraging deep neural networks to improve precision and generalization in image matting.

As previous research progressed from still images to videos, video matting emerged as a natural extension. However, collecting high-quality, per-frame alpha annotations for videos is expensive and time-consuming. Moreover, requiring users to provide auxiliary inputs (e.g., trimaps or masks) for every frame introduces substantial practical limitations. To mitigate these challenging issues, recent works \cite{video_deep1, video_deep2, one} have proposed auxiliary-free or single-frame-conditioned video matting techniques. However, without any strong priors, these methods often suffer from degraded performance due to the ambiguity of the matting task. 

To address these limitations, several methods \cite{modnet,RVM} have focused on human-centric scenarios, primarily targeting background removal in portrait images. These methods emphasize ease of use and accurate boundary reconstruction, particularly for challenging regions such as hair. 

While existing work has concentrated on background matting, we broaden the scope by targeting a wider range of occlusions, including hands, hair, transparent objects, and even semitransparent elements such as smoke or fire. These occlusions present unique challenges, such as intricate boundaries and the lack of clear trimap guidance. Moreover, since our proposed framework operates without any auxiliary inputs for test, conventional trimap- or mask-dependent approaches are inapplicable in real-world settings. In our trimap-free, real-world setting, the matting region must be inferred solely from the image content.

\subsection{Face Occlusion Segmentation}
Face occlusion segmentation is critical for a wide range of face-related tasks, such as face recognition \cite{occ_recog, Face_recognition_2}, face swapping \cite{occ_swap, face_swap2}, and facial reconstruction \cite{occ_recon, face_reconstruction_2}. Occlusions caused by external objects, such as hands, accessories, or hair, can significantly degrade the performance of these applications, making robust occlusion handling essential for real-world deployment.

To address this issues, several datasets~\cite{facedataset1, facedataset2} have been introduced, featuring occluded faces collected from in-the-wild scenarios. However, these datasets often suffer from key limitations, including limited scale, low resolution, and insufficient diversity in occlusion types. Furthermore, many lack high-quality segmentation annotations, or only provide coarse binary or categorical labels, which are inadequate for learning fine-grained occlusions.

\begin{figure}[t]
    \includegraphics[width=1.0\linewidth]{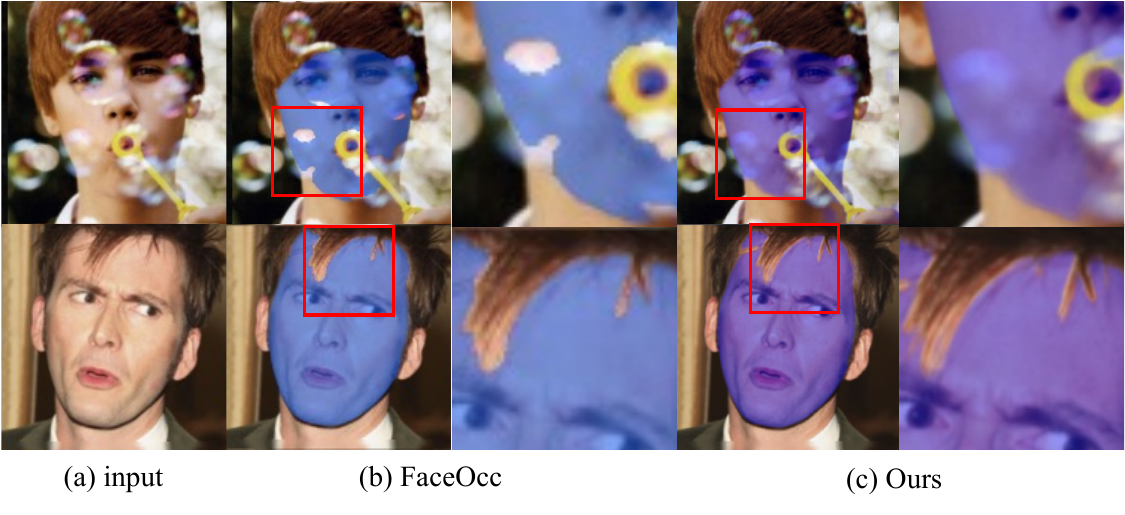}
    \vspace{-0.7cm}
    \caption{Comparison with FaceOcc \cite{faceocc}. (a) Input image. (b) FaceOcc fails to accurately separate hair from face due to hard binary masks. (c) Our method provides precise separation of facial and hair regions using alpha-based segmentation.}
    \Description{This figure shows an example output with ... (a brief textual description)}
\label{fig:limitation}
\vspace{-0.5cm}
\end{figure}

Recent works~\cite{natocc, faceocc} have attempted to address these challenges. However, as illustrated in Figure~\ref{fig:limitation}, segmentation-based methods are fundamentally limited in their ability to model soft transitions and semi-transparent occlusions, such as motion blur or translucent objects. These cases introduce ambiguities that cannot be effectively resolved with discrete label maps. As a result, relying solely on segmentation often leads to hard boundaries and visible artifacts in downstream applications, especially in video-based scenarios.

To overcome these limitations, advocate for a matting-based formulation. Notably, unlike segmentation, matting predicts continuous alpha values, enabling smooth foreground-background transitions and more accurate integration of occluded elements. This is particularly beneficial for face filtering and editing tasks, where natural compositing is crucial. 

In this work, we propose a novel framework for face occlusion matting that estimates high-quality alpha mattes for occluding regions. By replacing segmentation with matting, our method improves robustness to complex occlusion patterns and delivers more visually coherent results across various face-related applications.

\section{Problem Formulation} 

\subsection{Definition of Face Matting}
Face matting is the task of accurately extracting the facial region while separating occlusions that appear in front of the face. Unlike general image matting, where the target object is often ambiguous and requires auxiliary inputs such as trimaps or binary masks, face matting benefits from a well-defined foreground. Although auxiliary inputs provide explicit guidance, they impose additional computation and user costs. Instead, auxiliary-free face matting leverages domain-specific training to distinguish the foreground without requiring external input.

\begin{figure}
    \centerline{\includegraphics[width=1.0\linewidth]{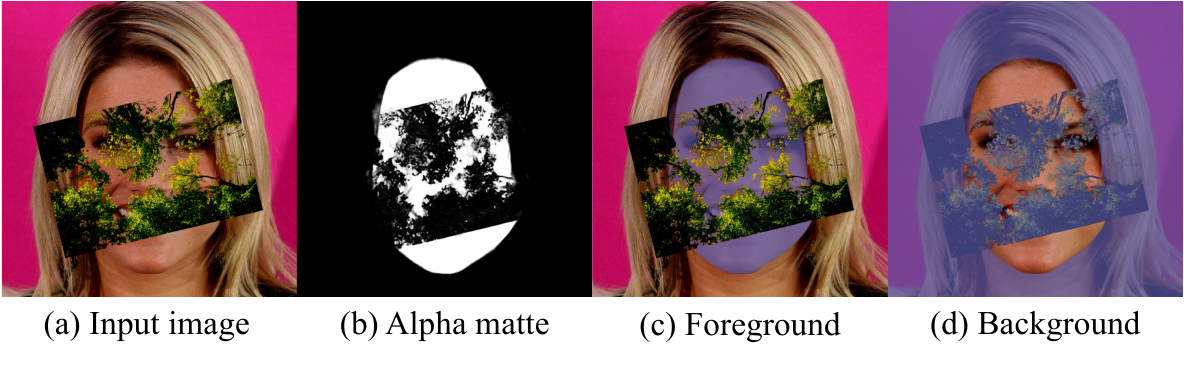}}
    \caption{Face matting definition in our framework. (a) Input image with partial occlusions. (b) Predicted alpha matte separating face and occluders. (c, d) Extracted foreground and background for compositing and manipulation.}
    \Description{}
\label{fig:def}
\end{figure}

In face matting, the foreground is initially defined as the skin area, encompassing all facial components, while the background consists of the remaining regions outside the foreground. In real-world scenarios, occlusions exist in front of the facial skin from the camera’s viewpoint, complicating the foreground-background distinction. Thus, while the skin is conceptually treated as the foreground, its physical placement in the scene contradicts this assumption. However, to maintain consistency with the conventional image matting formulation (\ref{eq:1}), we define the skin region as the foreground. Figure. \ref{fig:def} illustrates examples of input images, annotations, foreground, and background representations in the face matting task.

Performing face matting yields the alpha matte for the foreground, i.e. the skin, enabling flexible recomposition of either the skin or occlusions. In this study, occlusions are defined based on their feasibility in recomposing, specifically by determining which elements should be preserved after face-related tasks. Specifically, the skin region includes the face area with eyes, nose, and mouth, while the rest includes occlusions such as hair, ears, and hands. Non-body elements like heavy makeup are also considered occlusions, whereas purely transparent lenses without shadows or color are excluded. 



\subsection{Dataset Generation}
To construct a dataset tailored for face matting, both a high-quality face dataset and diverse occlusion sources are essential. We synthesized training samples by compositing occlusions onto face images under various conditions. This process results in CelebAMat, a new dataset designed to support realistic and occlusion-aware face matting. We newly introduce and release CelebAMat as a novel benchmark dataset for evaluating face matting models under diverse and challenging occlusion scenarios. We use CelebAMat both as the training dataset and as a benchmark for evaluating face matting performance under diverse occlusion conditions.

\begin{figure}
    \centerline{\includegraphics[width=1.0\linewidth]{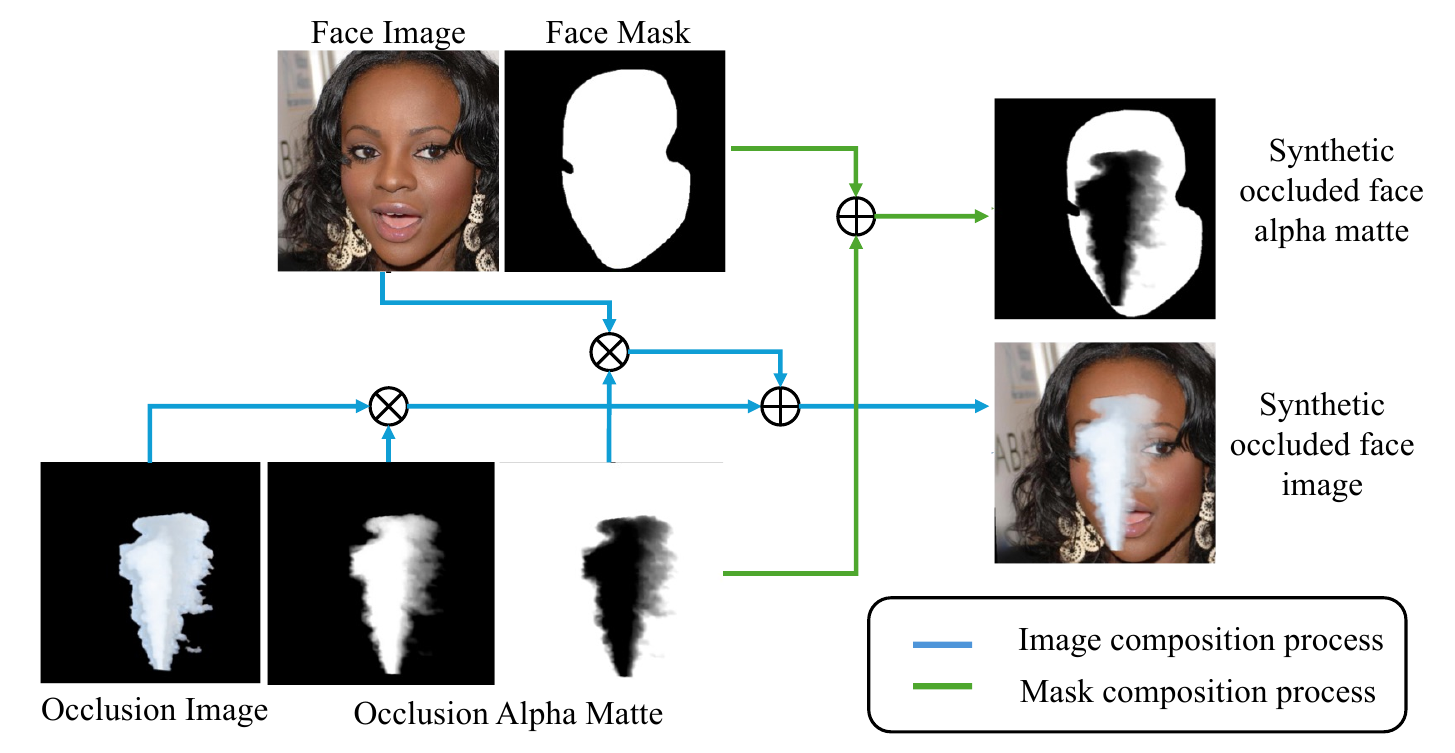}}
    \vspace{-0.25cm}
    \caption{Overview of face matting data generation. A clean face image and an occlusion image are composited to create a synthetic occluded face. The corresponding soft alpha matte is generated by blending the face mask and occlusion mask, enabling pixel-level soft mask ground truth for training. }
    \Description{This figure shows an example output with ... (a brief textual description)}
\label{fig:generation}
\vspace{-0.6cm}
\end{figure}

The CelebAMask-HQ dataset \cite{celeba} provides high-resolution face images with segmentation masks for facial attributes, which we employ for face matting. To ensure occlusion-free face images, samples with visible occlusions were removed from both training and testing sets, using the refined annotations and partitioning provided by \cite{natocc}. This results in a dataset of 24,602 training images and 716 test images, matching the original quantity. However, their annotations lack the precision required for matting applications and often fail to accurately capture fine-grained facial details such as glasses and facial hair.

For occlusion modeling, we utilize several dataset from the image matting datasets: SIMD \cite{semantic}, AM2k \cite{bridging}, hand segmentation dataset: HIU-data \cite{hiu}, and Describe Textures Dataset (DTD) \cite{dtd} for occlusion diversity. To enhance the realism of the dataset, we applied gaussian blur to the boundary regions of the HIU mask. 

Fig~\ref{fig:generation} illustrates the overall data generation process. For each dataset, occlusions are sampled strictly from the original training split during training, and from the test split during evaluation. During training, occlusion instances are inserted with random variations in size, orientation, and position to increase robustness. In contrast, during evaluation, we adopt a fixed set of occlusion configurations to form a consistent benchmark setting. 





\begin{figure*}[!]
    \centering
    \includegraphics[width=\textwidth]{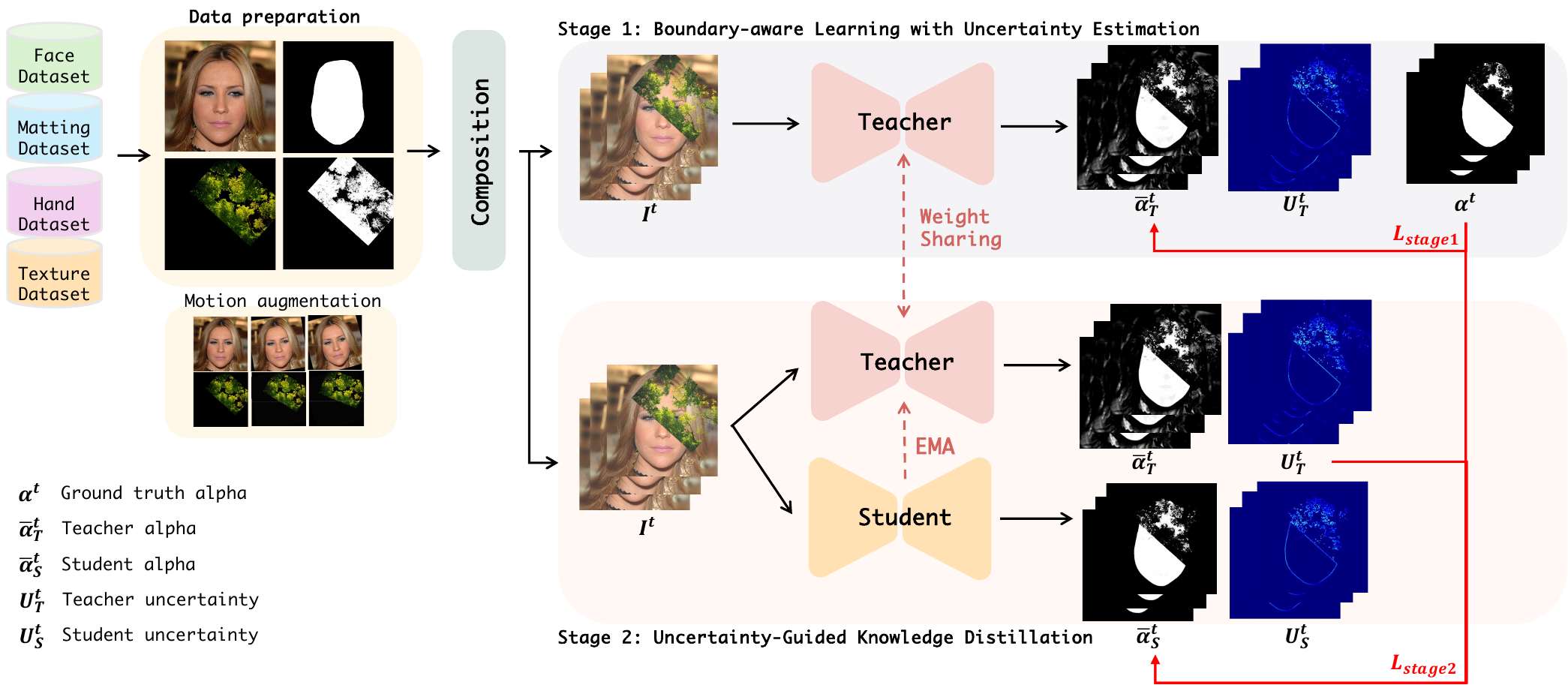}
    \caption{Overview of the full FaceMat pipeline. Multiple datasets, including face, hand, texture, and matting datasets, are combined through motion-aware composition to generate occlusion-rich training data. The framework then proceeds in two stages: Stage 1 trains a teacher model with boundary-aware learning and uncertainty estimation using trimaps; Stage 2 distills this knowledge to a student model via uncertainty-guided supervision, enabling trimap-free face matting under occlusions.}
    \Description{}
    \label{fig:framework}
\end{figure*}


\section{Methodology}

An overview of our proposed training framework \textbf{FaceMat}, is illustrated in Fig. \ref{fig:framework}. Conventional matting methods rely heavily on trimaps as spatial priors and loss weighting, with evaluation restricted to ambiguous regions. This often overlooks the semantic consistency of the full facial structure. 

To address this limitation, we propose a novel two-stage training strategy: 1) \textbf{Boundary-aware Learning with Uncertainty Estimation}: In the first stage, a teacher network is jointly trained to predict both the alpha matte and associated uncertainty map. The teacher model is trained using trimaps to focus on the precise separation of facial boundaries, like other conventional matting pipeline \cite{transmatting, mgmatting,ppmatting, danmatting, gcamatting, hattmatting, wildmatting, diffumatting, matteformer}.
2) \textbf{Uncertainty-Guided Knowledge Distillation (UGKD)}: In the second stage, the estimated uncertainty map is utilized to drive a locally adaptive distillation strategy that guides the corresponding student model. 

Uncertainty estimation has been increasingly adopted in deep learning to improve model robustness and interpretability \cite{uncertainty1, Uncertainty2, uncertainty3, uncertainty_devision}. It has proven effective in various computer vision tasks, including segmentation \cite{uncertainty_seg}, image classification \cite{uncertainty_classification}, and object detection \cite{uncertainty_detection}, where it is primarily used to quantify prediction confidence and refine model outputs. Unlike prior approaches that primarily utilize uncertainty as a measure of confidence, we propose a novel use of uncertainty as a supervisory signal. Specifically, we leverage the estimated uncertainty to guide the student model's learning in a fine-grained and spatially adaptive manner, which is crucial for modeling the soft transitions and boundary ambiguities inherent in image matting.  

While the previous matting models that apply uniform supervision across the entire image rely heavily on handcrafted trimaps, our framework adaptively focuses on boundary-critical regions where prediction uncertainty is highest. 

This mechanism encourages the student model to concentrate on boundary-critical regions, where alpha estimation exhibits the highest degree of uncertainty and thus requires more fine-grained supervision. As a result, the model not only learns to preserve comprehensive semantic representations across the entire image but also retains high fidelity in capturing intricate boundary details.


\subsection{Boundary-aware Learning with Uncertainty Estimation}

The matting task inherently involves a high degree of uncertainty, as it requires the prediction of alpha values that lie between the foreground and background from a visually mixed composition. This blending is influenced not only by color information but also by factors such as lighting direction, sensor characteristics, and scene dynamics. Compared to other vision tasks, matting exhibits greater ambiguity, particularly near object boundaries. Since these uncertain regions vary depending on local image conditions, a modeling approach that captures heteroscedasticity is essential for robust performance.


In our framework, the model is trained to jointly predict the alpha matte and a corresponding pixel-wise uncertainty map. This is achieved via an objective based on the NLL loss, which directly encourages the network to estimate a per-pixel variance conditioned on local image features. Consequently, the model implicitly learns to represent heteroscedastic aleatoric uncertainty, enabling it to allocate greater attention to ambiguous regions while maintaining robustness in more confident areas.

Furthermore, we apply the NLL formulation not only to the uncertainty map but also to the predicted alpha matte itself. This contributes to the spatial smoothness of the alpha map, yielding more natural transitions between regions. Additionally, it acts as an implicit regularizer for the uncertainty estimation, mitigating overconfidence and promoting stable learning dynamics.

\begin{equation} \label{eq:nll-nuc} 
    \mathcal{L}_{\text{NLL}-u}^{\beta} = \frac{1}{2} \left( \frac{(u - \mu_u)^2}{\sigma_u^2} + \log \sigma_u^2 \right) \cdot (\sigma_u^2)^{\beta}
\end{equation}

\begin{equation} \label{eq:nll-alpha} 
    \mathcal{L}_{\text{NLL}-\alpha}^{\beta} = \frac{1}{2} \left( \frac{(\alpha - \mu_\alpha)^2}{\sigma_\alpha^2} + \log \sigma_\alpha^2 \right) \cdot (\sigma_\alpha^2)^{\beta}
\end{equation}

In the first stage, we aim to capture fine-grained boundary details of the alpha matte. To this end, we utilize trimaps generated from the ground truth alpha to spatially constrain the loss computation. We adopt RVM \cite{RVM} as our baseline model, which is specifically designed for video matting, and apply all losses across the full temporal range of frames, i.e., for all \( t \in [1, T] \).

Following RVM, we employ the L1 regression loss \( \mathcal{L}_{\text{L1}} \), the pyramid Laplacian loss \( \mathcal{L}_{\text{lap}} \), and the temporal consistency loss \( \mathcal{L}_{\text{tc}} \). In addition, we incorporate negative log-likelihood (NLL) losses for both the predicted uncertainty map \( \mathcal{L}_{\text{NLL}-u}^{\beta} \) and the alpha matte \( \mathcal{L}_{\text{NLL}-\alpha}^{\beta} \). Note that all losses except for the NLL losses are masked by the unknown region in the trimap to emphasize learning in the ambiguous regions. The final objective function for the teacher model is defined as:
\[
\mathcal{L}_{\text{stage1}} = \mathcal{L}_{\text{L1}} + \mathcal{L}_{\text{lap}} + \mathcal{L}_{\text{tc}} + \mathcal{L}_{\text{NLL}-u}^{\beta} + \mathcal{L}_{\text{NLL}-\alpha}^{\beta}.
\]



\subsection{Uncertainty-guided Knowledge Distillation}
\indent
Conventional image matting models often  rely on auxiliary inputs such as trimaps to guide the model by providing coarse annotations of foreground, background, and unknown regions. These auxiliary cues are commonly used not only to inform the model duroing inference but also to constrain the loss computation by masking the supervision to ambiguous regions. 
In contrast, our approach is trimap-free, meaning that no such auxiliary hints are available during training of inference. Consequently, the model must learn semantic priors and structural details directly from the data. This can introduce a key challenge: the inherent trade-off between capturing high-level semantics and preserving fine-grained boundary details. Effectively balancing this trade-off is crucial for generating high-quality alpha mattes in unconstrained settings. 

To address this, we propose an uncertainty-guided knowledge distillation (UGKD) framework that adaptively modulates the supervision strength based on the estimated teacher model's confidence, or uncertainty. Specifically, since we have access to both the ground truth and the teacher's predictions, we leverage the teacher's uncertainty map to identify regions where the predictions are ambiguous or less reliable.

\begin{table*}[t]
\centering
\caption{Quatitative comparison of matting methods on the CelebAMat benchmark under various occlusion types. We report MSE and SAD scores (lower is better) across four test sets: SIMD, AM2k, HIU, and Random (Rand). This table serves as a baseline reference for evaluating performance under divese occlusion. Best and second-best results are marked in \textbf{bold} and \underline{underlined}, respectively. }
\resizebox{\textwidth}{!}{ 
\begin{tabular}{ccc | cc cc | cc cc}
\toprule
\multicolumn{3}{c|}{\textbf{Model Configuration}} & \multicolumn{4}{c|}{\textbf{Matting Test Dataset}} & \multicolumn{4}{c}{\textbf{Segmentation Test Dataset}} \\
\cmidrule(lr){1-3} \cmidrule(lr){4-7} \cmidrule(lr){8-11}
\raisebox{-2ex}{Network} & \raisebox{-2ex}{Auxiliary Input} & \raisebox{-2ex}{Encoder Type} & \multicolumn{2}{c}{SIMD} & \multicolumn{2}{c|}{AM2k} & \multicolumn{2}{c}{HIU} & \multicolumn{2}{c}{Rand} \\
& & & MSE($\downarrow$) & SAD($\downarrow$) & MSE($\downarrow$) & SAD($\downarrow$) & MSE($\downarrow$) & SAD($\downarrow$) & MSE($\downarrow$) & SAD($\downarrow$) \\
\midrule
UNet & -- & ResNet18 & 0.0693 & 31.2129 & 0.0454 & 17.1462 & 0.0578 & 20.2689 & 0.0651 & 25.5367 \\
Aematter\cite{aematter} & Trimap & Transformer & 0.0942 & 37.6412 & 0.0472 & 16.6684 & 0.0719 & 24.0475 & 0.0484 & 17.2676 \\
MGMatting\cite{mgmatting} & Mask & ResNet34 & 0.0645 & 29.3725 & 0.0360 & 14.2273 & 0.0469 & 17.2082 & 0.0319 & 13.1362 \\
MODNet\cite{modnet} & -- & MobileNetV2 & \underline{0.0457} & \underline{23.1507} & \underline{0.0311} & \underline{11.2826} & \underline{0.0266} & \textbf{9.6049} & \underline{0.0350} & \underline{12.3250} \\
RVM\cite{RVM} & -- & MobileNetV3 & \textbf{0.0301} & \textbf{20.0812} & \textbf{0.0105} & \textbf{5.5623} & \textbf{0.0199} & \underline{10.1634} & \textbf{0.0123} & \textbf{6.2677} \\
\bottomrule
\end{tabular}
}
\label{tab:benchmark}
\end{table*}

{\small
\begin{table*}[!]
  \caption{Quantitative comparison across different training settings on CelebAMat. Trimap(\checkmark) indicates that a trimap is used as part of the loss function. Each metric is reported as mean $\pm$ standard deviation (std). Best and second-best results are marked in bold and \underline{UNDERLINED}, respectively. The lowest std for each metric is shown in \textcolor{red}{red} text.}
  \label{tab:trimap_results}
  \centering
  \begin{tabular}{cccccccc}
    \toprule
    Setting & Trimap & MSE($\downarrow$) & SAD($\downarrow$) & Grad($\downarrow$) & Conn($\downarrow$) & IoU($\uparrow$) & Accuracy($\uparrow$) \\
    \midrule
    \multicolumn{8}{l}{Stage 1: Comparison of NLL-based Multi-Task Learning Variants} \\
    \midrule
    RVM\cite{RVM} & \checkmark 
      & 0.0202
      & 10.2900
      & \underline{0.0385 }
      & \underline{7.0865 }
      & 0.8249
      & 0.9528 \\
    Stage1 (NLL) & \checkmark
      & \textbf{0.0169 $\pm$ 0.00077 }
      & \underline{9.25 $\pm$ 0.3400 }
      & 0.0437 $\pm$ 0.0014
      & 13.3785 $\pm$ 2.0844
      & 0.6017 $\pm$ 0.0998 
      & 0.8180 $\pm$ 0.0797 \\
      
    \midrule
    \midrule
    \multicolumn{8}{l}{Stage 2: Comparison of Uncertainty-guided Knowledge Distillation} \\
    \midrule
    Stage1 Extended	& \checkmark
      & 0.0183 $\pm$ 0.00203 
      & 9.82 $\pm$ 1.8800 
      & 0.0437$\pm$ 0.0006
      & 12.6937 $\pm$ 1.620
      & 0.6333 $\pm$ 0.1340 
      & 0.8424 $\pm$ 0.0891 \\
      
    Stage1 Extended & 
      & 0.0236 $\pm$ 0.00041 
      & 16.38 $\pm$ 8.9800
      & \textbf{0.0379 $\pm$ 0.0018}
      & 7.6577 $\pm$ 1.5681
      & \textbf{0.8579 \textcolor{red}{$\pm$ 0.0036}}
      & \textbf{0.9602 \textcolor{red}{$\pm$ 0.0017}}\\
      
    Stage2 (UGKD) & 
      & \underline{0.0182 \textcolor{red}{$\pm$ 0.00028}}
      & \textbf{8.43 \textcolor{red}{$\pm$ 0.1600}}
      & 0.0424 \textcolor{red}{$\pm$ 0.0004}
      & \textbf{6.5764 \textcolor{red}{$\pm$ 0.4824}} 
      & \underline{0.8408  $\pm$ 0.0181 }
      & \underline{0.9529  $\pm$ 0.0067 }\\
    \bottomrule
  \end{tabular}
\end{table*}
}

To enforce stronger learning in these regions, we introduce an uncertainty-weighted $L1$ regression loss between the ground truth and the student model’s predicted alpha matte. This encourages the student model to focus more on semantically complex or visually uncertain areas during training. The objective of stage 2 is defined as $L_{\text{stage2}} = L_{l1}^{\text{soft}} + L_{\text{lap}}$. Here, $L_{l1}^{\text{soft}}$ is the uncertainty-weighted $L1$ regression loss between the predicted alpha matte and the ground truth:

\begin{equation}
L_{l1}^{\text{soft}} = \left\| w_{\text{unc}} \odot (\alpha - \alpha_{\text{gt}}) \right\|_1
\end{equation}

where $\alpha$ denotes the predicted alpha matte by the student model, $\odot$ is element-wise multiplication, $\alpha_{\text{gt}}$ is the ground truth, and $w_{\text{unc}}$ is the uncertainty-based spatial weight map derived from the teacher model's uncertainty. 


To emphasize regions where the teacher model is less confident, we define the uncertainty-based weighting map $w_{\text{unc}}$ as a linear function of the predicted variance $\sigma_u^{\text{teacher}}$: 
\[
w_{\text{unc}} = w_1 + w_2 \cdot \sigma_u^{\text{teacher}}
\]
where $w_1 = 2$ and $w_2 = 2$. This formulation ensures that higher uncertainty leads to greater loss contribution, thereby encouraging the student model to prioritize learning from difficult of ambiguous regions. 
To improve training stability, the teacher model is updated throughout the training process using an Exponential Moving Average (EMA) of the student model parameters.




\section{Experiments}

\subsection{Benchmarking}
\indent
To benchmark our proposed face matting dataset, CelebAMat, we evaluate five representative matting models:
U-Net with a ResNet-18 encoder pretrained on ImageNet; AEMatter \cite{aematter}, a transformer-based model that utilizes trimaps as auxiliary input; MGMatting \cite{mgmatting}, a mask-guided approach with a ResNet-34 backbone; and two trimap-free video matting models, MODNet\cite{modnet} and RVM \cite{RVM}.

Although methods with auxiliary inputs are generally expected to generalize well, we observe a significant performance degradation when applied to CelebAMat without additional adaptation. This phenomenon reflects the auxiliary input inconsistency problem, where a mismatch between training and inference distribution of auxiliary inputs undermines generalization. To ensure fair and consistent evaluation, we trained all models on CelebAMat using the same training settings.

Quantitative results are reported in Table \ref{tab:benchmark}. Among all evaluated models, RVM demonstrates the best overall performance across diverse occlusion scenarios. Accordingly, we adopt RVM as the baseline in our framework and extend it with our proposed uncertainty-guided distillation strategy to further enhance matting performance under various occlusions. 





\subsection{Implementation Details}
For the occlusion segmentation datasets, HIU~\cite{hiu} and DTD~\cite{dtd}, which are significantly larger than occlusion matting datasets such as SIMD~\cite{semantic} and AM2k~\cite{bridging}, we randomly selected 200 samples from each dataset for training to ensure a balanced comparison. As our method only utilize still image datasets, we synthesize motion across adjacent frames by applying affine transformations, resizing, horizontal flipping, color jittering, and random pauses to occluding objects, following the approach of RVM \cite{RVM}. 


We evaluate our model on the CelebAMat test dataset using four standard matting metrics: Mean Squared Error (MSE), Sum of Absolute Differences (SAD), Gradient error (Grad), and Connectivity error (Conn), all computed within the unknown region of the trimap. Additionally to measure the model's ability to capture semantic object understanding at the image level, we report binary segmentation metrics: Intersection-over-Union (IoU) and pixel-wise accuracy. To further evaluate robustness in real-world scenarios, we conduct additional experiments on the RealOcc~\cite{natocc} which is face occlusion segmentation dataset, reporting IoU, accuracy, and recall as complementary evaluation metrics.

\begin{table}[!]
    \centering
    \caption{Quantitative comparison on RealOcc~\cite{natocc} in terms of IoU, accuracy, and recall between RVM and our method.}
    \begin{tabular}{cccc}
    \toprule
    \textbf{Setting} & \textbf{IoU} & \textbf{Accuracy} & \textbf{Recall} \\
    \midrule
    RVM\cite{RVM}   & 0.4099 & 0.8432 & 0.4876 \\
    Ours & 0.7121 & 0.9197 & 0.9084 \\
    \bottomrule
    \end{tabular}
    \label{tab:quantitative_realocc}
    \vspace{-0.5cm}
\end{table}

\begin{figure*}[!]
    \centering
    \includegraphics[width=\textwidth]{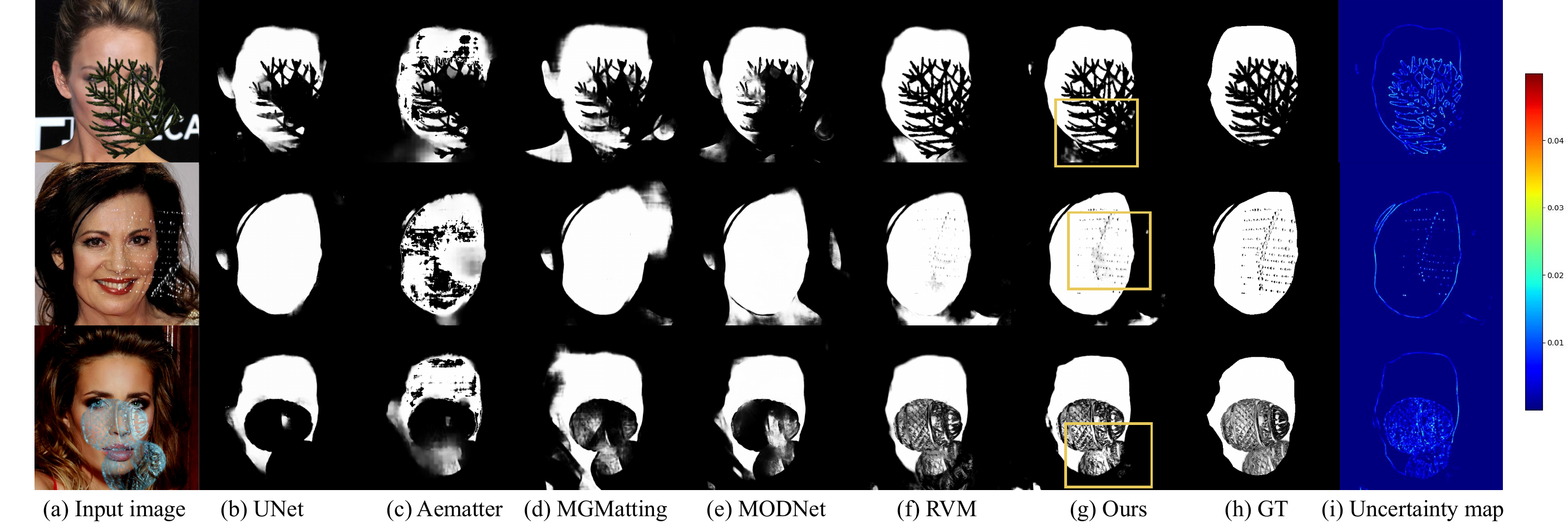}
    \caption{Qualitative comparison on CelebAMat benchmark. (a) shows the occluded input image. As highlighted in the yellow boxes, previous methods (b-f) struggle to preserve fine facial boundaries under occlusions. In contrast, our proposed method (g) predicts a sharper and more accurate alpha matte. (i) presents the uncertainty map estimated by the teacher model in Stage 1, which guides the student model to focus on ambiguous regions and enhances boundary accuracy during distillation. }
    \Description{}
    \label{fig:qualitative}
\end{figure*}

\subsection{Quantitative Comparison}

Quantitative results for our FaceMat framework on the CelebAMat benchmark are summarized in Table \ref{tab:trimap_results}.

In stage 1, we apply negative log-likelihood (NLL) loss independently to both the predicted uncertainty map and the alpha matte. This dual supervision leads to performance improvements by enabling the model not only to estimate accurate alpha values but also to self-assess the confidence of its predictions. Moreover, this setup promotes implicit multitask learning, which has regularizing effects and helps mitigate overfitting.

To extend the model's capacity beyond boundary refinement, we compare several training strategies that progressively build toward full-scene semantic understanding. In particular, we investigate: (1) extended training of the Stage 1 model, (2) training without using trimaps in the objective, and (3) full Stage 2 training with uncertainty-guided knowledge distillation (UGKD). 

We observe that relying on trimaps in Stage 1 improves local boundary quality but limits global generalization, especially in semantic metrics such as IoU and pixel accuracy. Removing the trimap yields better semantic performance but often degrades accuracy in matting-specific metrics like MSE, SAD. In contrast, our full Stage 2 model with UGKD achieves a strong balance between the two, maintaining competitive boundary precision while improving global semantic consistency. 

The results demonstrate that the proposed two-stage framework effectively preserves fine-grained boundaries while learning to reason over semantically meaningful structures. This outcomes is largely attributed to the use of pixel-wise uncertainty, which provides soft, spatially adaptive guidance, which cannot flexibly account for varying degrees of ambiguity across the image.

Furthermore, we also observe that UGKD reduces intra-model variance, called as epistemic uncertainty, leading to more stable and consistent alpha predictions. This is especially beneficial in high-frequency or semi-transparent regions, where even minor inconsistencies may result in perceptual artifacts.

To further assess the generalizability of our method, we also evaluate it on the RealOcc. The results are summarized in Table \ref{tab:quantitative_realocc}.

\subsection{Qualitative Comparison}
Figure \ref{fig:qualitative} shows qualitative comparisons between our FaceMat and several state-of-the-art matting models on the CelebAMaat benchmark. Each example includes an occluded input image (a), followed by the predicted alpha mattes from previous methods (b-f) our result (g), and the teacher-predicted uncertainty map (i). 

As highlighted in the yellow boxes, previous methods exhibit difficulties in accurately separating facial structures from occluding elements, especially in regions such as hair boundaries, ears. These models often produce over-smoothed or overly hard transitions, resulting in perceptual artifacts.

In contrast, our method (g) delivers more precise alpha mattes, preserving fine-grained facial boundaries while excluding occluders. Notably, the uncertainty map (i) reveals that the teacher model assigns higher uncertainty to ambiguous regions such as transition zones skin and hair. The spatial distribution aligns closely with the unknown regions of conventional trimap and demonstrates that the uncertainty map effectively serves as a soft attention prior durinig distillation.

These results highlights the benefit of our uncertainty-guided distillation strategy, which enables the student model to concentrate supervision on visually ambiguous areas, leading to more coherent and robust alpha matte predictions under real-world occlusion conditions. 

As shown in Figure~\ref{fig:qualitative}, the predicted uncertainty map exhibits high activation near facial boundaries and occlusion edges, which align well with the unknown region of the trimap. This observation supports the idea that the uncertainty map has been effectively trained to reflect inherently ambiguous regions, such as the transition zones between foreground and background. 

\begin{figure}
    \centerline{\includegraphics[width=1.0\linewidth]{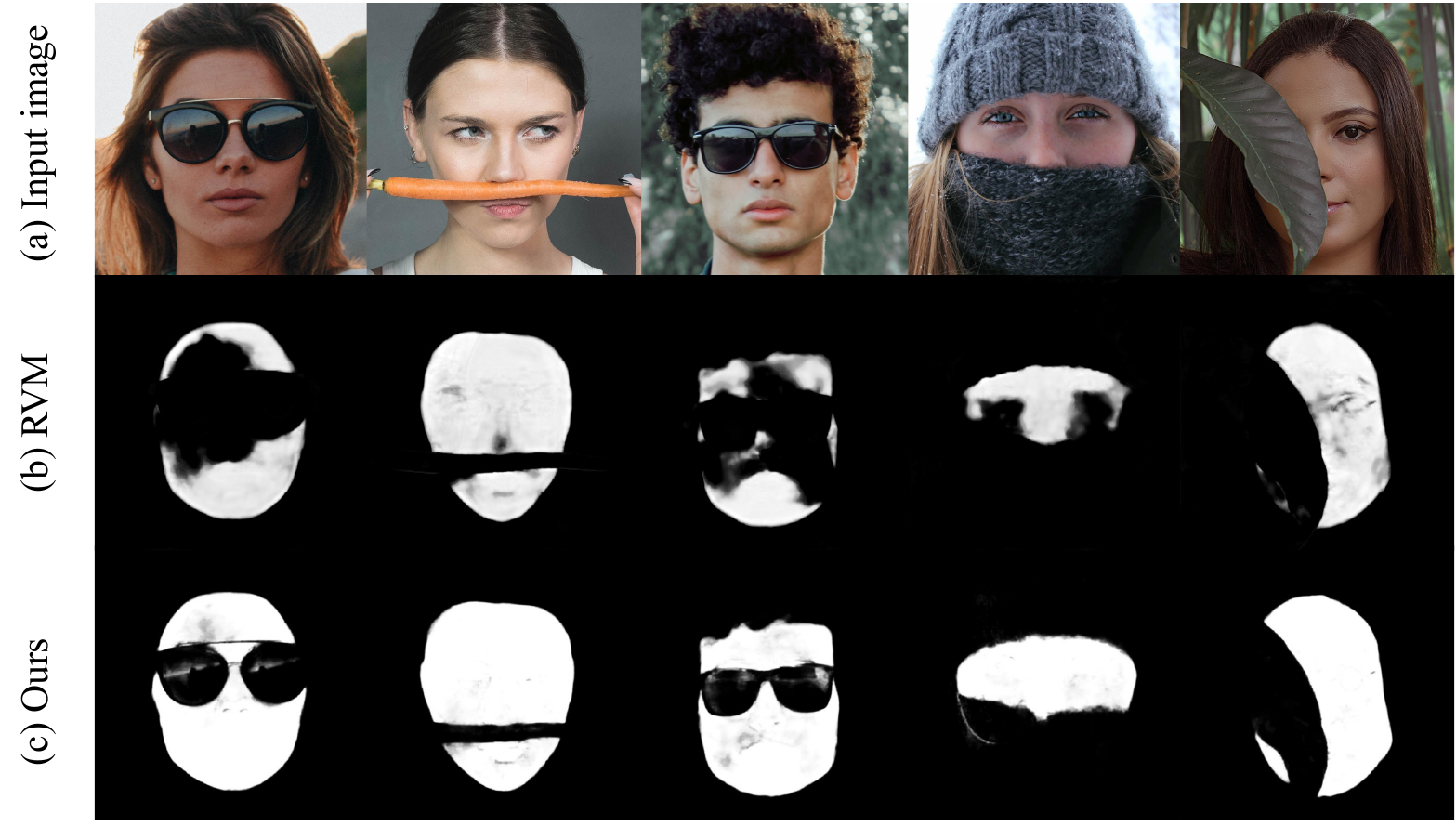}}
    \caption{Qualitative results on the RealOcc dataset. (a) shows an input image with natural occlusions. (b) illustrates a failure case where the baseline struggles with complex, in-the-wild appearance. In contrast, (c) successfully separates the face region from heavy occlusions.}
    \Description{}
\label{fig:realocc}
\end{figure}

Figure \ref{fig:realocc} illustrates qualitative comparison on RealOcc. Compared to CelebAMat, which belongs to the same domain as the training data, our method exhibits improved generalization to unseen domains. UGKD facilitates the learning of structural representations that are crucial for domain adaptation. 

{\scriptsize
\renewcommand{\arraystretch}{0.8}
\setlength{\tabcolsep}{1.2pt}
\begin{table}[!]
\centering
\caption{Ablation study on occlusion dataset composition in ResNet18}
\vspace{-0.25cm}
\label{tab:abl_comp}
    {\scriptsize 
    \begin{tabular}{cccc|cc|cc|cc|cc}
        \toprule
        \multicolumn{4}{c|}{\multirow{2}{*}{\textbf{Occlusion Train Dataset}}} & \multicolumn{4}{c|}{\textbf{Matting Test Dataset}} & \multicolumn{4}{c}{\textbf{Segmentation Test Dataset}} \\
        \cmidrule(lr){5-12}
        \multicolumn{4}{c|}{} & \multicolumn{2}{c|}{SIMD} & \multicolumn{2}{c|}{AM2K} & \multicolumn{2}{c|}{HIU} & \multicolumn{2}{c}{Rand} \\
        \midrule
        SIMD & HIU & Rand & AM2K & MSE ($\downarrow$) & SAD ($\downarrow$) & MSE ($\downarrow$) & SAD ($\downarrow$) & MSE ($\downarrow$) & SAD ($\downarrow$) & MSE ($\downarrow$) & SAD ($\downarrow$) \\
        \midrule
        \checkmark &  &  &  & 0.1248 & 51.7981 & 0.1337 & 47.1108 & 0.1213 & 42.7249 & 0.1581 & 53.6633 \\
        \checkmark & \checkmark  &  &  & 0.0806 & 35.9724 & 0.0703 & 25.5113 & 0.0738 & 25.8287 & 0.0947 & 31.9772 \\
        \checkmark  & \checkmark  & \checkmark  &  & 0.0871 & 40.2320 & 0.0820 & 31.1358 & 0.0793 & 30.0110 & 0.1015 & 37.0328 \\
        \checkmark  & \checkmark  & \checkmark  & \checkmark  & \textbf{0.0693} & \textbf{31.2129} & \textbf{0.0454} & \textbf{17.1462} & \textbf{0.0578} & \textbf{20.2689} & \textbf{0.0651} & \textbf{25.5367} \\
        \bottomrule
    \end{tabular}
    }
\end{table}
}

{\scriptsize
\renewcommand{\arraystretch}{0.8}
\setlength{\tabcolsep}{1.75pt}
\begin{table}[!ht]
\centering
\caption{Ablation study on fixed occlusion ratio in the training dataset for ResNet18}
\vspace{-0.25cm}
\label{tab:ratio_fixed}
    {\footnotesize 
    \begin{tabular}{c|cc|cc|cc|cc}
        \toprule
        \multirow{3}{*}{\raisebox{-6ex}{\shortstack[c]{Fixed\\Occlusion\\Ratio}}}
        & \multicolumn{4}{c|}{\textbf{Matting Test Dataset}} 
        & \multicolumn{4}{c}{\textbf{Segmentation Test Dataset}} \\
        \cmidrule(lr){2-9}
        & \multicolumn{2}{c|}{SIMD} & \multicolumn{2}{c|}{AM2K} & \multicolumn{2}{c|}{HIU} & \multicolumn{2}{c}{Rand} \\
        \cmidrule(lr){2-9}
        & MSE ($\downarrow$) & SAD ($\downarrow$) 
        & MSE ($\downarrow$) & SAD ($\downarrow$) 
        & MSE ($\downarrow$) & SAD ($\downarrow$) 
        & MSE ($\downarrow$) & SAD ($\downarrow$) \\
        \midrule
        1     & 0.3258 & 99.3865 & 0.4201 & 112.3609 & 0.4445 & 117.2768 & 0.4297 & 117.7057 \\
        0.75  & 0.1343 & 52.8224 & 0.1478 & 49.9889  & 0.1210 & 41.7132  & 0.1740 & 56.4217  \\
        0.5   & 0.1232 & 49.1341 & 0.1151 & 39.7154  & 0.1139 & 38.2213  & 0.1314 & 44.1395  \\
        0.25  & \textbf{0.0693} & \textbf{31.2129} & \textbf{0.0454} & \textbf{17.1462} 
               & \textbf{0.0578} & \textbf{20.2689} & \textbf{0.0651} & \textbf{25.5367} \\
        \bottomrule
    \end{tabular}
    }
\end{table}
}

{\scriptsize
\renewcommand{\arraystretch}{0.8}
\setlength{\tabcolsep}{1.75pt}
\begin{table}[!]
\centering
\caption{Ablation study on occlusion ratio scheduling patterns in the training dataset for RVM.  
Pattern 1 increases the ratio by 0.2 every 10 epochs: [0.1, 0.3, 0.5, 0.7, 0.9, 1.0].  
Pattern 2 increases it by 0.1 every 10 epochs: [0.1, 0.2, 0.3, 0.4, 0.5, 0.6].}
\vspace{-0.25cm}
\label{tab:ratio_pattern}
    {\footnotesize 
    \begin{tabular}{c|cc|cc|cc|cc}
        \toprule
        \multirow{3}{*}{\raisebox{-5ex}{\shortstack[c]{Occlusion\\pattern}}}
        & \multicolumn{4}{c|}{\textbf{Matting Test Dataset}} 
        & \multicolumn{4}{c}{\textbf{Segmentation Test Dataset}} \\
        \cmidrule(lr){2-9}
        & \multicolumn{2}{c|}{SIMD} & \multicolumn{2}{c|}{AM2K} & \multicolumn{2}{c|}{HIU} & \multicolumn{2}{c}{Rand} \\
        \cmidrule(lr){2-9}
        & MSE ($\downarrow$) & SAD ($\downarrow$) 
        & MSE ($\downarrow$) & SAD ($\downarrow$) 
        & MSE ($\downarrow$) & SAD ($\downarrow$) 
        & MSE ($\downarrow$) & SAD ($\downarrow$) \\
        \midrule
        fixed 0.25 & \textbf{0.0301} & \textbf{20.0812} & \textbf{0.0105} & \textbf{5.5623} & \textbf{0.0199} & \textbf{10.1634} & \textbf{0.0123} & \textbf{6.2677} \\
        pattern 1   & 0.0318 & 22.3034 & 0.0187 & 8.4664 & 0.0206 & 11.0830 & 0.0184 & 8.2502 \\
        pattern 2   & 0.0303 & 21.4591 & 0.0176 & 7.8937 & 0.0206 & 10.7454 & 0.0190 & 7.9013 \\
        \bottomrule
    \end{tabular}
    }
\end{table}
}

\subsection{Ablation study and Analysis}
To demonstrate the effectiveness of occlusion diversity and the impact of the occlusion ratio in the CelebAMat dataset, we conduct an ablation study using a simple U-Net architecture. As shown in Table~\ref{tab:abl_comp}, training with the full set of occlusion datasets improves performance, highlighting the benefit of incorporating diverse occlusion types. In addition, Table~\ref{tab:ratio_fixed} indicates that an occlusion ratio of 0.25 yields the best results, enabling the model to better learn facial boundaries and components under partially occluded conditions.

Building on these findings, Table~\ref{tab:ratio_pattern} presents an additional ablation study using the RVM framework to investigate scheduled occlusion ratio increases during training. Although various patterns were explored, the fixed 0.25 ratio consistently achieved the best performance.
We hypothesize that maintaining a moderate level of occlusion encourages the model to focus on learning high-level semantic structures of the face, rather than overfitting to heavily occluded cases.



\subsection{Real-world Application}
To demonstrate the generality and practical potential of our occlusion-aware face matting model, we introduce an application for face filters under occluded conditions. As illustrated in Figure~\ref{fig:teaser}, our method enables stable face filter rendering even when key facial attributes are partially or fully occluded. 

\section{Conclusion}
We introduce a novel task, face matting, aimed at enabling occlusion-aware face transformation in real-world scenarios. To support this task, we construct CelebAMat, a large-scale benchmark dataset synthesized from clean facial images and diverse occluders with realistic motion augmentations. Leveraging this dataset, we propose FaceMat, a two-stage learning framework that accurately predicts alpha mattes under occlusions by integrating boundary-aware learning and uncertainty-guided knowledge distillation. 

Our extensive experiments demonstrate that our FaceMat not only preserves fine-grained facial boundaries but also improves robustness under challenging visual conditions, such as hands, accessories, or motion blur. We also validate its practical utility by applying it to downstream video tasks, where existing methods often fail due to occlusion artifacts or lack of semantic understanding. 

Finally, as shown in \textit{Supplementary material}, ensuring temporal consistency in facial inpainting remains an open challenge, pointing to a valuable direction for future work beyond alpha matte prediction. 



\clearpage

\balance
\bibliographystyle{ACM-Reference-Format}
\bibliography{bib_tex}

\clearpage
\appendix
\input{supplementary.tex}

\end{document}

%% file: supplementary.tex
\settopmatter{printacmref=false} 
\definecolor{pastelblue}{RGB}{204,229,255}
\definecolor{pastelgreen}{RGB}{204,255,229}
\definecolor{pastelyellow}{RGB}{255,255,204}
\definecolor{pastelpink}{RGB}{255,204,229}
\definecolor{pastelpurple}{RGB}{229,204,255}



\copyrightyear{2025}
\acmYear{2025}
\setcopyright{acmlicensed}
\setcctype{by-nc-nd}
\acmConference[MM '25]{Proceedings of the 33rd ACM International Conference on Multimedia}{October 27--31, 2025}{Dublin, Ireland}
\acmBooktitle{Proceedings of the 33rd ACM International Conference on Multimedia (MM '25), October 27--31, 2025, Dublin, Ireland}
\acmDOI{10.1145/3746027.3755060}
\acmISBN{979-8-4007-2035-2/2025/10}

\acmSubmissionID{1885}


\title{Supplementary : Uncertainty-Guided Face Matting for Occlusion-Aware Face Transformation}

\author{Hyebin Cho}
\affiliation{%
  \institution{Korea Advanced Institute of Science \& Technology} 
  \department{School of Electrical Engineering}
  \city{Daejeon}
  \country{Republic of Korea}}
\email{hyebin.cho@kaist.ac.kr} 


\author{Jaehyup Lee}
\authornote{Corresponding author}
\affiliation{%
  \institution{Kyungpook National University}
    \department{School of Computer Science and Engineering}
  \city{Daegu}
  \country{Republic of Korea}}
\email{jaehyuplee@knu.ac.kr}

\renewcommand{\shortauthors}{Cho et al.}


\maketitle


\indent
This supplementary material provides additional results and implementation details for our paper "Uncertainty-Guided Face Matting for Occlusion-Aware Face Transformation". It includes further details about dataset generation, stage comparison, additional restuls and a video demo of the proposed method.

\section{Dataset Generation}


Face matting aims to extract facial regions from both background and occluding objects in front of the face. Following the conventional matting formulation, we define the face as the foreground, while background elements and occlusions are treated as the background. Table~\ref{tab:def_for} compares our definition of foreground and background with those used in related tasks.

For facial data, CelebAMask-HQ dataset~\cite{celeba} provides rough skin attribute masks but includes certain occluding elements, such as glasses. To address this, \cite{natocc} introduced CelebAMask-HQ-WO, a variant in which all occlusions are removed. We also adopt this version in our framework to obtain clean foreground reference.

The definition of ``occlusion'' can vary across tasks. Previous works on face occlusion segmentation~\cite{natocc, faceocc} adopt task-specific definitions. In our case, we define occlusion based on application-driven considerations. Since our goal is to support face-centric applications such as facial editing and realistic video compositing, we define occlusion in terms of its semantic role and its impact on facial visibility in the final output. Table \ref{tab:def_occ} summarizes our classification of occluding and non-occluding elements based on this principle. 

Following our definitions of foreground and background, we constructed CelebAMat dataset accordingly. Table \ref{tab:datasets} summarizes the partitions of each dataset used in our pipeline. Each dataset provides pixel-wise annotations in the form of binary masks or alpha mattes. 
Using these annotations, we constructed our synthetic CelebAMat dataset designed for the face matting task.

\begin{figure*}[!]
    \centering
    \includegraphics[width=\textwidth]{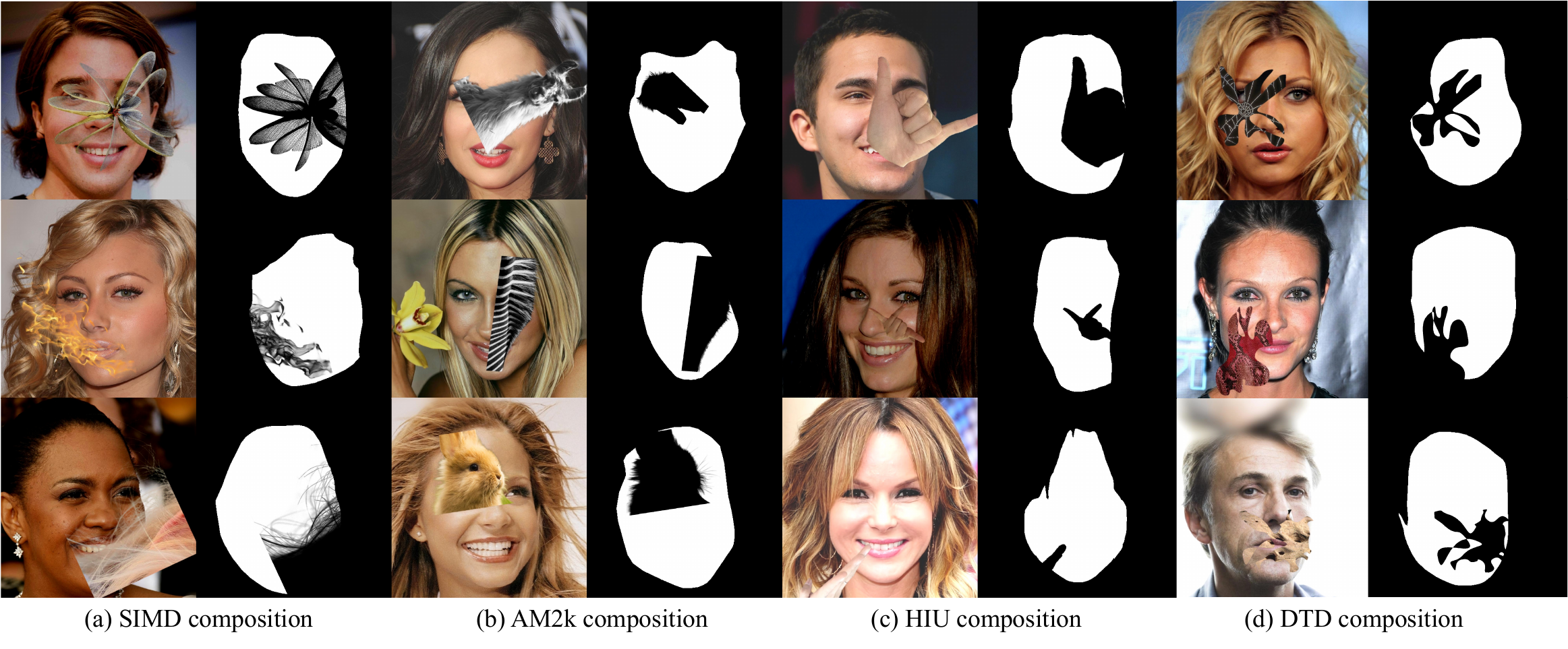}
    \caption{Examples from our CelebAMat test dataset. Subfigures (a-d) show sample compositions using different occlusion datasets: (a) SIMD, (b) AM2k, (c) HIU, and (d) synthetic random shapes. All images are augmented with resizing, rotation, horizontal flipping, and color jittering. }
    \Description{}
    \label{fig:datageneration_example}
\end{figure*}

Figure~\ref{fig:datageneration_example} presents sample image compositions from each occlusion dataset. All examples were augmented with random horizontal flipping, resizing, rotation, and color jittering. Gaussian blur is applied only to the HIU-based composition in (c). In (d), random shapes are synthesized by connecting randomly sampled points with curves.

\begin{table}
    \centering
    \caption{Definition of Foreground}
    \begin{tabular}{c | c}
    \toprule
    Task & Foreground \\
    \midrule
    Matting with auxiliary input & Defined by auxiliary input \\
    Portrait matting & Human \\
    Animal matting & Animal \\
    \textbf{Face matting (Ours)} & Skin including face components \\
    \bottomrule
    \end{tabular}
    \label{tab:def_for}
\end{table}

\begin{table}
  \caption{Definitions of Occlusions}
  \label{tab:def_occ}
  \centering
  {\footnotesize 
  \begin{tabular}{c|c|>{\centering\arraybackslash}p{2.5cm}|>{\centering\arraybackslash}p{2.5cm}}
    \toprule
    Paper & Task & Occlusion & Non-occlusion \\
    \midrule
    \cite{faceocc} & Segmentation & Deep shadow, Strong makeup, Tattoo, Tongue, Eyeglass frame, Color lens, Mirror reflection & Colorless transparent lens, Beard, Scalp \\
    \hline
    \cite{natocc} & Segmentation & Sunglass, Ear, Eyeglass including the colorless transparent lens & Shadow, Strong makeup, Tattoo, Beard, Scalp\\
    \hline
    \textbf{Ours} & Matting & Beard, Color lens, Ear, Eyeglass frame & Colorless transparent lens, Shadow, Strong makeup, Tattoo, Scalp\\
    \bottomrule
  \end{tabular}
  }
\end{table}

\begin{table}[!ht]
\caption{Partition of each datasets}
\begin{center}
\begin{tabular} {@{\extracolsep{6pt}}c|cccc}
    \toprule
    dataset & face & occlusion & train & test \\
    \midrule
    CelebAMask-HQ-WO & \checkmark &  &24602 & 716 \\
    \midrule
     SIMD & &\checkmark &720 & 78 \\
    AM2k & &\checkmark & 1800 & 200 \\
     HIU & &\checkmark & 8804 & 2204 \\
     DTD & &\checkmark & 4512 & 1128\\
    \midrule
    CelebAMat(Ours) & \checkmark &\checkmark&24602 & 716 \\
    \bottomrule
\end{tabular}
\end{center}
\label{tab:datasets}
\end{table}

\begin{figure}[!ht]
    \includegraphics[width=1.0\linewidth]{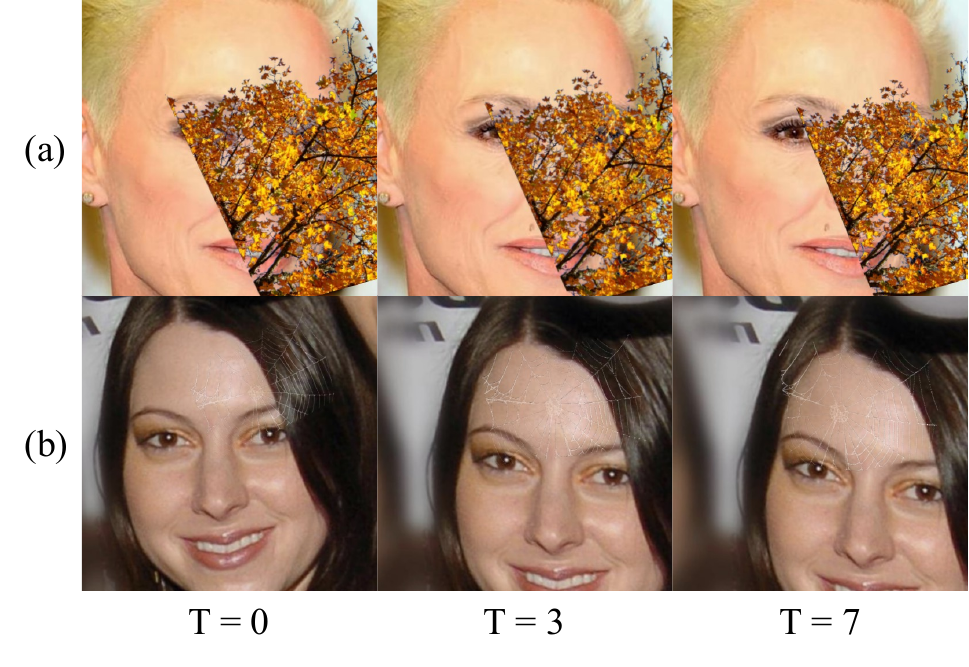}
    \caption{Motion augmentation for CelebAMat. (a) shows motion applied to the occluding objects, while (b) shows motion applied to the face region. Motion is simulated by interpolating between two random affine transformations across time steps, creating smooth synthetic sequences from still images. }
    \Description{}
    \label{fig:motion}
\end{figure}

Inspired by the recurrent decoder and temporal consistency loss in RVM \cite{RVM}, we simulate motion directly within our still-image dataset using affine transformations. For each image, two random affine matrices are generated and interpolated across the training sequence length. This produces smooth synthetic motion across frames without requiring actual video input. To preserve annotation quality, the motion magnitude is adaptively scaled based on sequence duration. Representative examples are shown in Figure~\ref{fig:motion}.

\section{Stage Comparison}

\begin{figure*}[!]
    \centering
    \includegraphics[width=\textwidth]{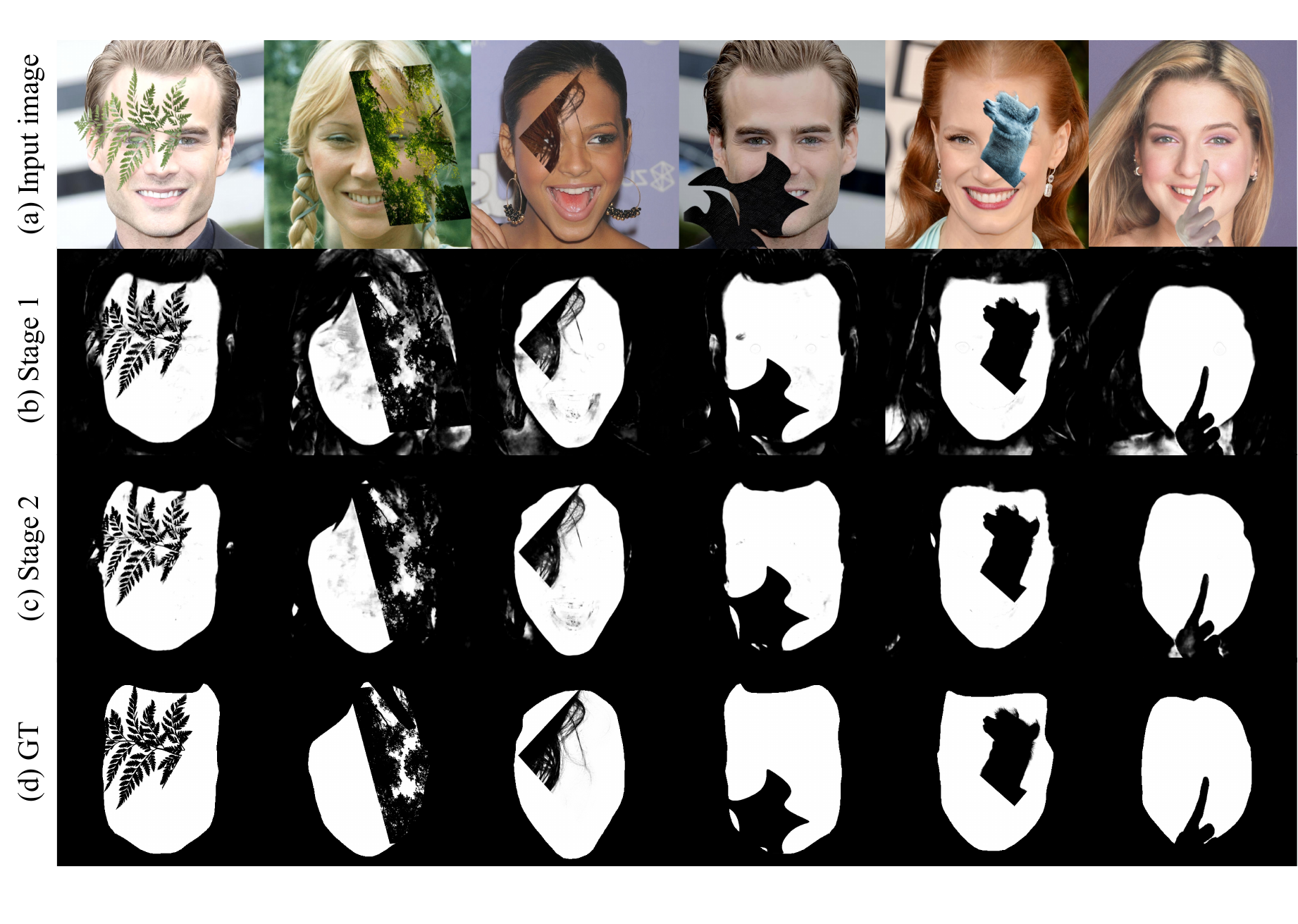}
    \caption{Qualitative results on CelebAMat. (a) shows the occluded input image. (b) depicts the predicted alpha matte from our our method, which preserves final facial boundaries and excludes occluders. (c) presents the uncertainty map estimated by the teacher model in Stage 1, which highlights ambiguous regions and guides the student model during distillation, improving boundary accuracy. 
    }
    \Description{}
    \label{fig:qualitative_stage}
\end{figure*}

Our proposed FaceMat framework adopts a two-stage training pipeline. Figure~\ref{fig:qualitative_stage} illustrates the qualitative results of each stage, allowing a clearer analysis of their respective contributions. 

Figure~\ref{fig:qualitative_stage} (b) shows the result from Stage 1, where the teacher model is trained to jointly predict both the alpha matte and the uncertainty map. In Figure~\ref{fig:qualitative_stage} (c), the Stage 2 output demonstrates notable improvements, particularly driven by the uncertainty-guided supervision. The model exhibits reduced confusion in background regions while maintaining sharper boundary predictions, especially in challenging areas such as hair and occlusions. 

\section{Additional Qualitative Results}

\begin{figure*}[!]
    \centering
    \includegraphics[width=\textwidth]{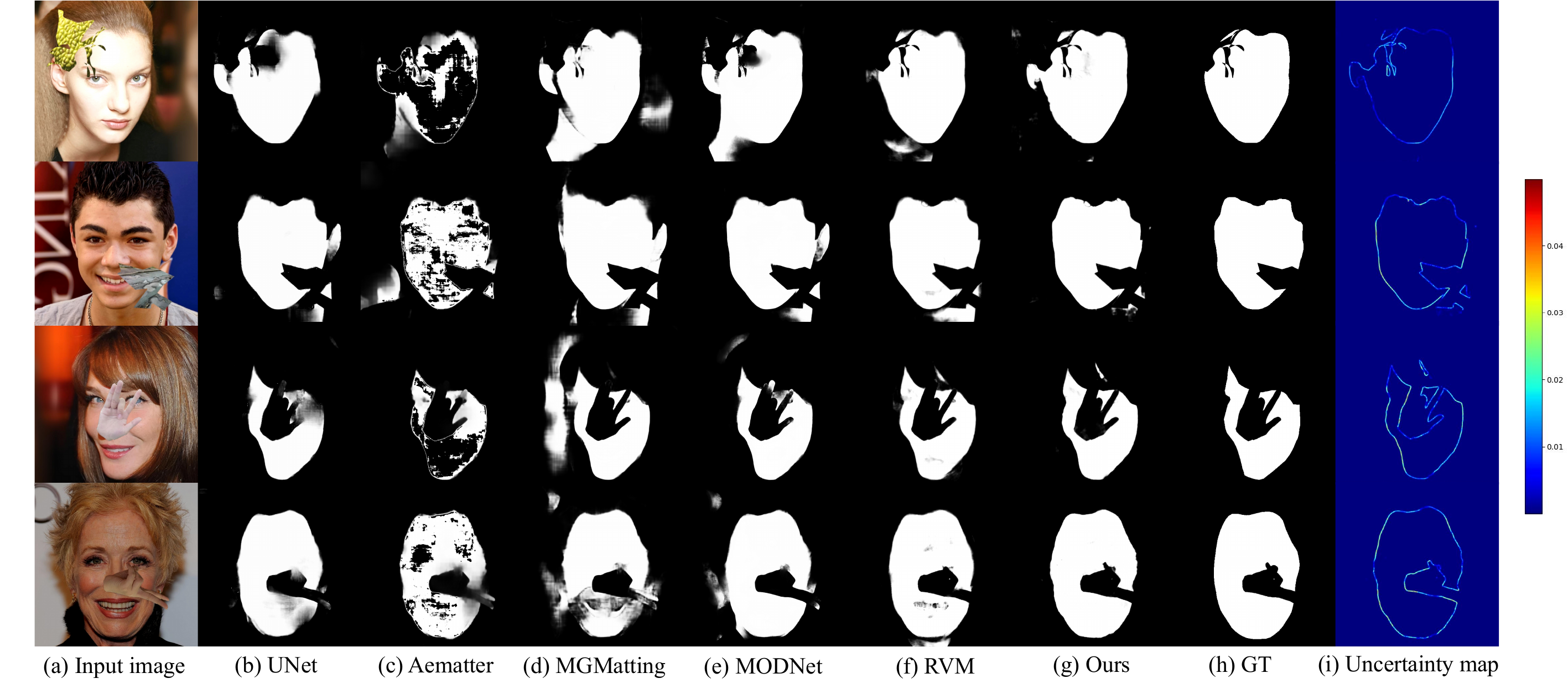}
    \caption{Additional qualitative comparison on CelebAMat benchmark. (a) shows the occluded input image. As highlighted in the yellow boxes, previous methods (b-f) struggle to preserve fine facial boundaries under occlusions. In contrast, our proposed method (g) predicts a sharper and more accurate alpha matte. (i) presents the uncertainty map estimated by the teacher model in Stage 1, which guides the student model to focus on ambiguous regions and enhances boundary accuracy during distillation. }
    \Description{}
    \label{fig:qualitative_celeba}
\end{figure*}

We present additional qualitative results on CelebAMat benchmark in figure~\ref{fig:qualitative_celeba}. 

To further assess real-world generalization, we evaluate our model on RealOcc~\cite{natocc}, a dataset designed for benchmarking face occlusion segmentation in natural scenes. Since our training data consists of synthetically composed occluded faces, it is important to validate performance in unconstrained, real-world environments.

However, RealOcc \cite{natocc} includes some challenging samples with extremely heavy occlusions, often leaving only a minimal portion of the skin visible. Moreover, the dataset is primarily tailored for segmentation tasks and focuses on common occlusion types such as sunglasses, hands, and face masks. 

\begin{figure*}[!]
    \centering
    \includegraphics[width=\textwidth]{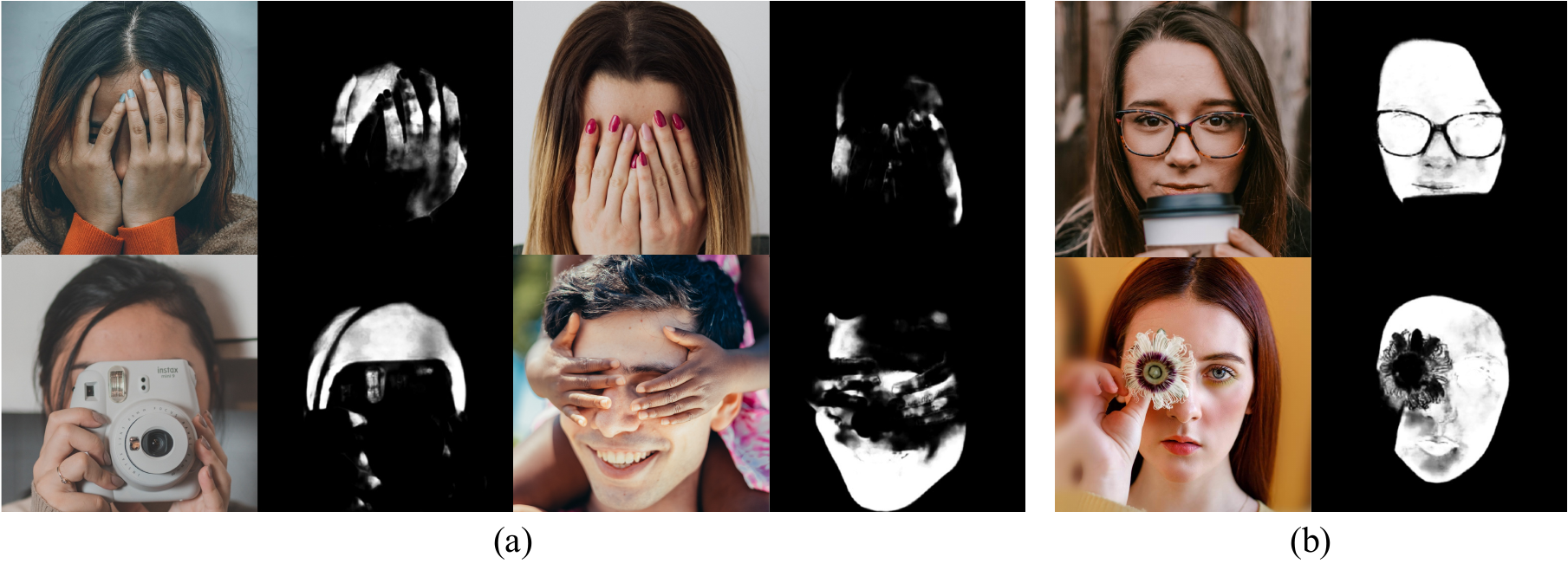}
    \caption{Failure cases in RealOcc dataset. (a) Heavy occlusion leading to inaccurate segmentation. (b) Facial parts (e.g., nose, lips) incorrectly preserved in the alpha matte.}
    \Description{}
    \label{fig:failure}
\end{figure*}

Figure~\ref{fig:failure} highlights typical failure cases of our method. When the face entirely covered by shadows or no facial features are visible, the model often struggles to localize the facial regain. In some cases, the predicted alpha matte includes unintended facial components, leading to inaccurate foreground estimation.

\begin{figure*}[!]
    \centering
    \includegraphics[width=\textwidth]{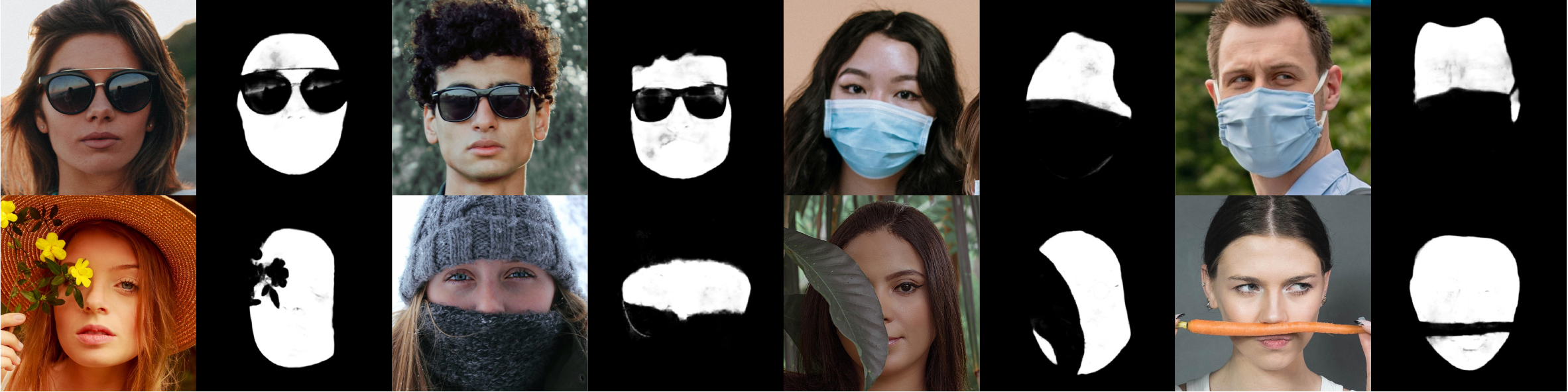}
    \caption{More output examples generated on RealOcc, showing performance under real-world occlusions.}
    \Description{}
    \label{fig:realocc}
\end{figure*}

\section{Ablation Study}
We compare different uncertainty weighting strategies during training: a linear function, an exponential function, and a linear function without Exponential Moving Average (EMA) updates for the student model. In the exponential formulation, the uncertainty weight is defined as: 

\[
w_{\text{unc}} = exp(w \cdot \sigma_u^{\text{teacher}})
\]

where $\sigma_u^{\text{teacher}}$ denotes the uncertainty predicted by the teacher model, $w$ is empirically set to 2. For EMA-based student updates, we use a decay rate to 0.97.

As shown in Table~\ref{tab:ablation}, all configurations achieve comparable quantitative performance across standard matting metrics. However, we adopt the linear weighting with EMA strategy as our final configuration, as it consistently produces more visually stable and semantically coherent alpha mattes in qualitative comparisons.

\begin{table}[h]
    \begin{tabular}{lccccc}
    \toprule
    Setting & EMA & MSE($\downarrow$) & SAD($\downarrow$) & IoU($\uparrow$) & Accuracy($\uparrow$) \\
    \midrule
    Linear  &  \checkmark  & 8.5257 & 0.8297 & 0.9492 & 0.9835 \\
    Linear  &   & 8.4930 & 0.8320 & 0.9502 & 0.9810 \\
    Exp     & \checkmark   & 8.8442 & 0.8426 & 0.9540 & 0.9826 \\
    \bottomrule
    \end{tabular}
    \caption{Comparison of different training schedules and EMA usage on key evaluation metrics.}
    \label{tab:ablation}
\end{table}

\section{Application Examples}

We apply our FaceMat model to a real-world application to demonstrate the generality and potential of the face matting. Using popular face filters (e.g., from the Snow app by Naver), we compare results with and without applying the predicted alpha matte. 

\begin{figure*}[!]
    \centering
    \includegraphics[width=\textwidth]{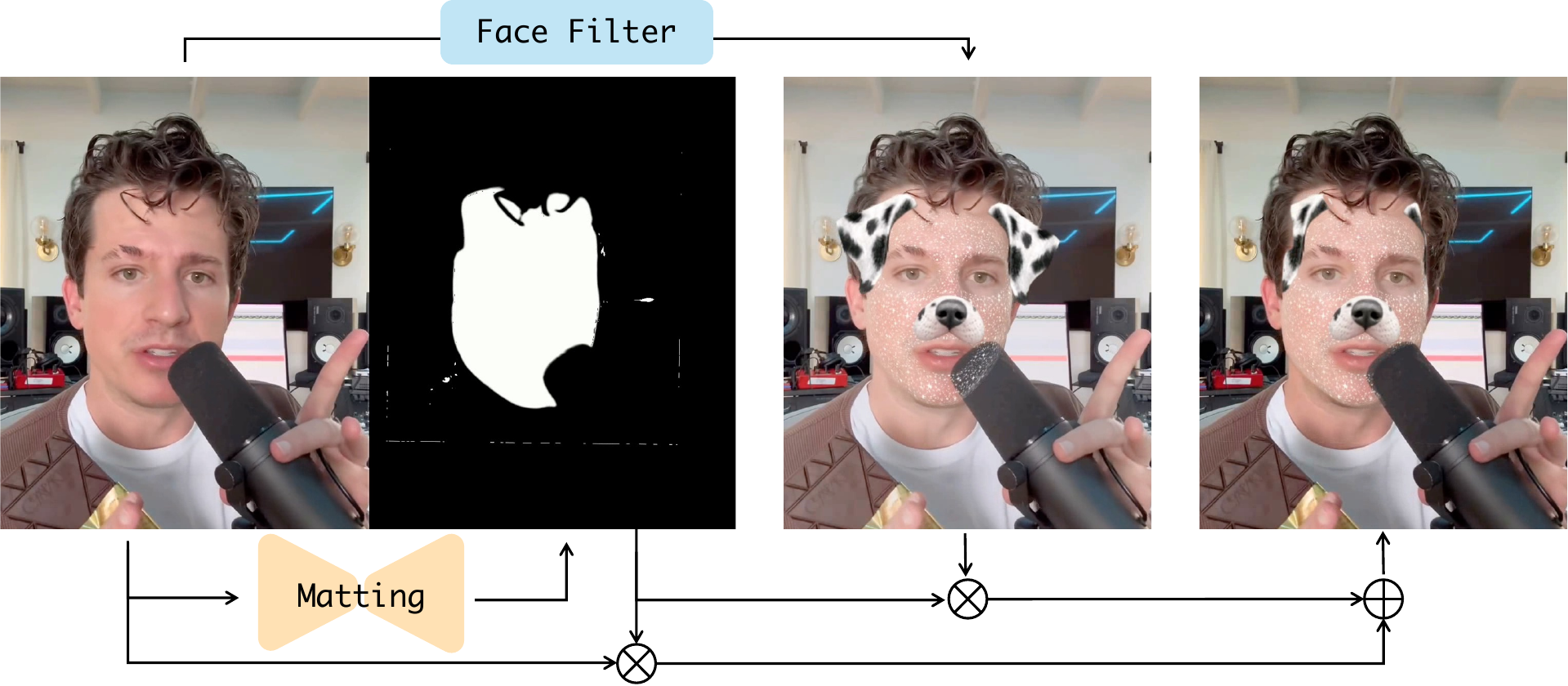}
    \caption{Visual pipeline of occlusion-aware face filter application using the predicted alpha matte.}
    \Description{}
    \label{fig:application_step}
\end{figure*}

Figure~\ref{fig:application_step} illustrates the process of generating a more natural and occlusion-aware face-filtered video. We first obtain an alpha matte for the face region using our model. Then, by recomposing the original face-filtered video with the matte, we ensure that the filter is applied only to visible facial regions, resulting in more realistic and semantically accurate outputs.

\begin{figure*}[!]
    \centering
    \includegraphics[width=\textwidth]{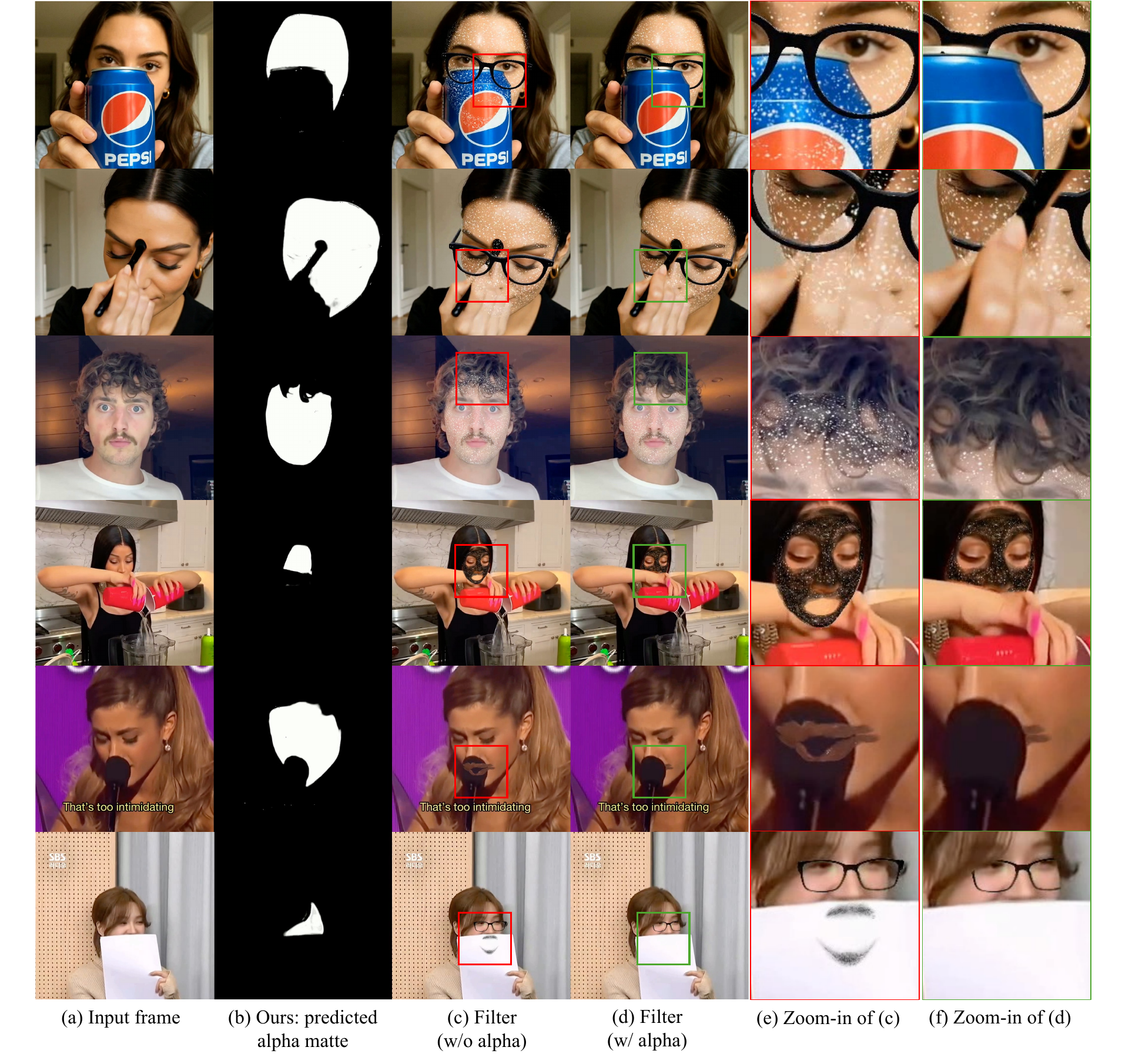}
    \caption{Additional examples of applying face matting to occlusion-aware face filtering. Each row shows (a) the occluded input image, (b) the predicted alpha matte by FaceMat, and (c–d) a comparison of filter application without and with matte-based compositing. Our approach consistently prevents filters from being applied to occluders (e.g., hair, microphones), resulting in more natural and semantically accurate outputs across various real-world scenarios.}
    \Description{}
    \label{fig:application_sample}
\end{figure*}

Figure~\ref{fig:application_step} illustrates the process of applying our predicted alpha matte to enhance face filter placement. We also experimented with inpainting occluded facial regions; however, achieving high-quality results remains difficult due to the lack of inpainting models that ensure temporal consistency across frames.

As shown in Figure~\ref{fig:application_sample}, when occlusions appear in front of the face, naively applying filters often results in unrealistic renderings, where the filter overlays occluding objects. By contrast, reapplying the filters with our predicted alpha matte restricts them to visible facial regions, resulting in more realistic and visually coherent outcomes.

